\documentclass[runningheads]{llncs}
\usepackage{graphicx}
\usepackage{comment}
\usepackage{amsmath,amssymb} % define this before the line numbering.
\usepackage{color}
\usepackage{booktabs}
\usepackage{threeparttable}
\usepackage{multirow}
\usepackage{color}
\usepackage{array}
\usepackage{wrapfig}
\usepackage[width=122mm,left=12mm,paperwidth=146mm,height=193mm,top=12mm,paperheight=217mm]{geometry}

\begin{document}
% \renewcommand\thelinenumber{\color[rgb]{0.2,0.5,0.8}\normalfont\sffamily\scriptsize\arabic{linenumber}\color[rgb]{0,0,0}}
% \renewcommand\makeLineNumber {\hss\thelinenumber\ \hspace{6mm} \rlap{\hskip\textwidth\ \hspace{6.5mm}\thelinenumber}}
% \linenumbers
\pagestyle{headings}
\mainmatter
\def\ECCVSubNumber{100}  % Insert your submission number here

\title{Dynamic Fusion Network For Light Field Depth Estimation} % Replace with your title
\author{Yongri Piao, Yukun Zhang, Miao Zhang\thanks{Corresponding Author},Xinxin Ji}
\institute{ Dalian University of Technology, China\\
        {\tt\small\{yrpiao, miaozhang\}@dlut.edu.cn}
        {\tt\small\{jxx0709,zhangyukun\}@mail.dlut.edu.cn}}

% INITIAL SUBMISSION
%\begin{comment}
%\titlerunning{ECCV-20 submission ID \ECCVSubNumber}
%\authorrunning{ECCV-20 submission ID \ECCVSubNumber}
%\author{Anonymous ECCV submission}
%\institute{Paper ID \ECCVSubNumber}
%\end{comment}
%******************

% CAMERA READY SUBMISSION
\begin{comment}
\titlerunning{Abbreviated paper title}
% If the paper title is too long for the running head, you can set
% an abbreviated paper title here
%
\author{First Author\inst{1}\orcidID{0000-1111-2222-3333} \and
Second Author\inst{2,3}\orcidID{1111-2222-3333-4444} \and
Third Author\inst{3}\orcidID{2222--3333-4444-5555}}
%
\authorrunning{F. Author et al.}
% First names are abbreviated in the running head.
% If there are more than two authors, 'et al.' is used.
%
\institute{Princeton University, Princeton NJ 08544, USA \and
Springer Heidelberg, Tiergartenstr. 17, 69121 Heidelberg, Germany
\email{lncs@springer.com}\\
\url{http://www.springer.com/gp/computer-science/lncs} \and
ABC Institute, Rupert-Karls-University Heidelberg, Heidelberg, Germany\\
\email{\{abc,lncs\}@uni-heidelberg.de}}
\end{comment}
%******************
\maketitle

\begin{abstract}
Focus-based methods have shown promising results for the task of depth estimation. However, most existing focus-based depth estimation approaches  depend on  maximal sharpness of the focal stack. Out-of-focus information in the focal stack poses challenges for this task. In this paper, we propose a dynamically multi-modal learning strategy which incorporates RGB data and the focal stack in our framework. Our goal is to deeply excavate the spatial correlation in the focal stack by designing the spatial-correlation perception module and dynamically fuse multi-modal information between RGB data and the focal stack in a adaptive way by designing the multi-modal dynamic fusion module. The success of our method is demonstrated by achieving the state-of-the-art performance on two datasets. Furthermore, we test our network on a set of different focused images generated by a smart-phone camera to prove that the proposed method not only broke the limitation of only using light field data, but also open a path toward practical applications of depth estimation on common consumer level cameras data. The code is available: https://github.com/OIPLab-DUT/Light-Field-for-Depth-Estimation.
\end{abstract}

\section{Introduction}

Depth estimation is a crucial step for understanding geometric relations within a scene. Accurate and reliable depth information plays an important role in computer vision including object tracking \cite{ghasemi2014scale,li2019adnet:}, scene understanding \cite{hazirbas2016fusenet}, virtual reality (VR) \cite{wozniak2019depth,Chih2017Rapid}, autonomous driving \cite{chen2015deepdriving} and pose estimation \cite{park2019multi,Lorenzo2017Depth}.
\begin{figure}
\begin{center}
\includegraphics[width=0.99\linewidth]{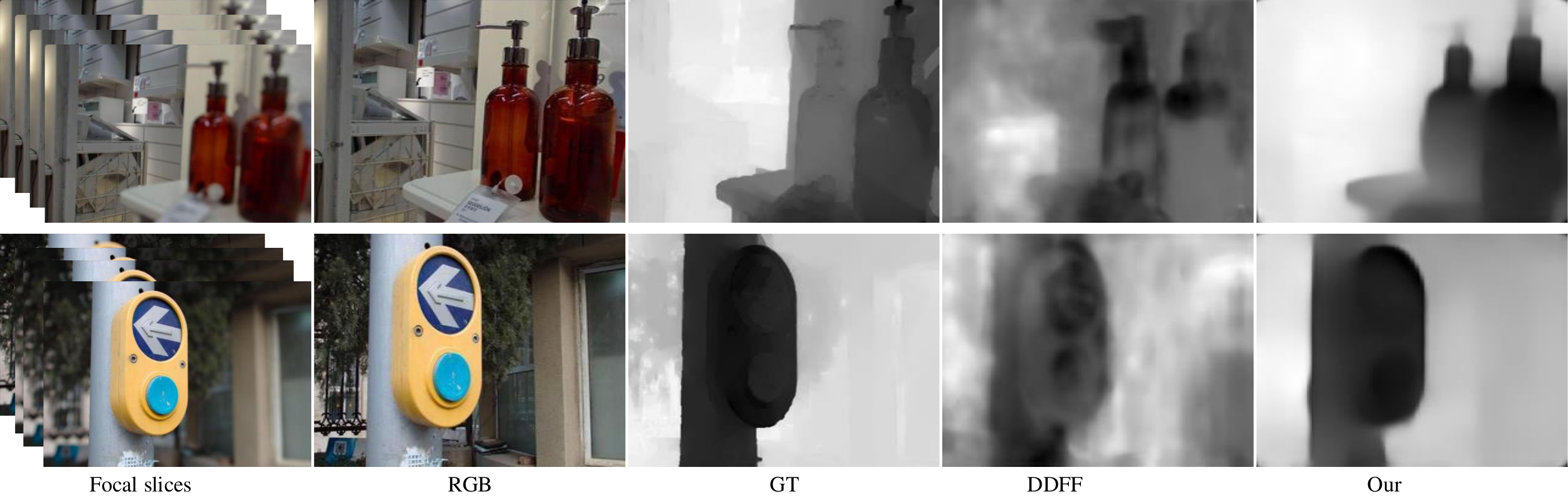}
\end{center}
\vspace{-4mm}
   \caption{From left to right: we show focal stack, RGB image, depth map, the result of DDFF, the result of our method in a scene.}
\label{fig:short1}
\vspace{-4mm}
\end{figure}
Depth estimation can be broadly classified into active and passive acquisition. In contrast to active techniques that involve sending a controlled energy beam and detecting the reflected energy \cite{liebe2004three,areann2001laser}, passive techniques are image-based methods and are more in accord with the human visual perception of the depth, i.e., humans use a great variety of vision-based passive depth cues such as texture, edges, size perspective, binocular disparity, motion parallax, occlusion effects and variations in shading. Monocular depth estimation \cite{srinivasan2018aperture,Chen2019Attention,mal2018sparse}, as low-cost, convenient and efficient passive techniques, has attracted lots of interest lately. However, depth estimation from a single image of a generic scene is an ill-posed problem, due to the inherent ambiguity of mapping an intensity or color measurement into a depth value. On the other hand, inspired by the analogy to human depth perception, multiview depth estimation has achieved great success, including binocular and multi-view stereo \cite{roberts2011structure,kendall2017end,pang2017cascade}. The similarities and correspondences produced between each pixel in the images produce far superior results. However, these approachs are sensitive to imaging systems and require careful alignment and calibration in the setup.

The light field enables a unique capability of post-capture refocusing. Fig.1 shows an example of the light field. A stack of focal slices are generated as they are taken at different depths, which contain abundant spatial parallax information. Furthermore, focusness information caters to human’s visual fixation that allows our eyes to maximize the focus we can give to the object in a scene. In spite of this promising characteristics for focusness information, there are a few studies documenting their efficacy for depth estimation. Early works mainly aim at determining the depth of a pixel by measuring its sharpness or focus at different images of the focal stack \cite{pertuz2013analysis,thelen2008improvements,mahmood2012nonlinear}. Later on, several approaches based on deep learning have been proposed \cite{hazirbas2018deep}. These methods use convolutional neural networks (CNNs) to extract effective focusness information for facilitating depth estimation instead of hand-crafted features.

While these methods demonstrate that focusness information is useful for depth estimation, there is still large room for further improvement in terms of two key aspects: (1) How we deeply excavate the spatial correlation between focal slices for obtaining useful focusness information is critical for depth estimation. Since  the different focal slices are focused at different depths,  the spatial correlation between the focal slices is closely  related to depth variation of the objects in a scene. Most of previous focal-based depth estimation networks used a standard 2D CNN to learns filters that extend across the entire focal stack. However, spatial correlation between the focal slices is likely to be ignored. As a result, the focusness information has not been well-captured. (2) How do we effectively fuse focusness features and RGB information to reduce information loss in the depth map? While focusness information in the focal slices provides implicit depth cues, leading to better depth estimation, out-of-focus areas with unknown sharpness could be prone to information loss error, leading to inaccurate depth estimation. As shown in Fig. 1, the result of DDFF which based on focal slices \cite{hazirbas2018deep} losses some detail and contains a lot of noise. Considering that the RGB image contains high quality sharpness and can be used to compensate for missing data in the out-of-focus area of the focal slices, we believe combining multi-modal information is beneficial to improve the accuracy of depth maps. Albeit most of recent methods performed data fusion by employing some manually set, such as, sum fusion, weighted fusion, concatenate fusion. Those methods are unable to take full advantages of multi-modal information between RGB images and focal slices.

Our core insight is that we can leverage RGB data and the focal stack to learn an estimation model of depth by deeply excavating the spatial correlation in the focal stack and fusing multi-modality cues between RGB image and focal slices. Concretely, our contributions are mainly three-fold:
\begin{itemize}
\item We proposed a spatial-correlation perception module for correlating the focus and depth. Based on the observe that different focal slices possess focus area of multiple scales and focused at different depth, a pyramid ConvGRU is designed to excavate the spatial correlation between different focal slices and sequently pass multi-scale focusness information  along the depth direction.
\item We propose a multi-modal dynamic fusion module (MDFM) in which multi-modalities features are fused in an adaptive manner. This fusion strategy allows the filter parameters to dynamically change with the input focusness features in the process of convolution with RGB features, thereby avoiding information  loss in the depth map.
 \item We demonstrate the effectiveness of the proposed model on two light field datasets \cite{li2014saliency}. The results show that our approach achieves superior performance over the state-of-the-art approaches. Moreover, we also validate our model on Mobile Phone dataset \cite{suwajanakorn2015depth} which contains focal slices captured by a smart-phone camera. This further shows a positive step towards practical application of our model with common consumer level cameras.
 \end{itemize}

 \section{Related Work}

The related works can be divided into two categories: light field depth estimation using traditional approaches and learning-based approaches.

\subsection{Traditional Approaches}
There is a wide range of methods for light field depth estimation. The field can be roughly divided into methods based on EPI analysis, multi-view stereo matching based approaches and focus-based approaches. \cite{zhang2016light} proposed a 4D light-field depth estimation method via epipolar plane image analysis and locally linear embedding. \cite{johannsen2016sparse} employed a sparse decomposition design to leverage the depth-orientation relationship on its epipolar plane images. Another popular approach is using total variation regularization. \cite{mahmood2013shape} proposed a nonlinear Total Variation (TV) based method for recovering 3D shape of an object by diffusing several initial depth maps obtained through different focus measures. \cite{moeller2015variational} proposed an non-convex minimization scheme to determine depth maps based on prior knowledge. \cite{javidnia2018application} applied a preconditioned alternating direction method of multipliers (PADMM) with a new cost function to generate a noise-free depth map. Jeon, et al. proposed an algorithm to estimate the multi-view stereo correspondences with sub-pixel accuracy using the cost volume \cite{ jeon2015accurate}. \cite{wang2015occlusion} developed an occlusion-aware depth estimation algorithm to deal with occlusion edges.

 Although the above traditional methods can produce good depth maps, these methods require presetting for the depth estimation with careful parameter tuning. Furthermore, they tend to be difficult to generalize into all scenes.

\subsection{Learning-based Approaches}
Recently, deep learning, in particular Convolutional Neural Networks (CNNs), has successfully broken the bottleneck of traditional methods in a wide range of fields. Such as, super-resolution  \cite{yoon2017light}, novel view generation \cite{kalantari2016learning} and material recognition \cite{wang20164d}. For the depth estimation, \cite{shin2018epinet} introduced a fully-convolutional neural network for highly accurate depth estimation. \cite{anwar2017depth} predicted depth from a single  focal slice using deep neural networks by exploiting dense overlapping patches. \cite{hazirbas2018deep} suggested the first end-to-end learning method to compute depth maps from the focal stack. \cite{srinivasan2018aperture} estimated scene depths from a single image, using the information provided by a camera’s aperture as supervision. They introduced two differentiable aperture rendering functions that use the input image and predicted depths to simulate the depth-of-field effects caused by real camera apertures. Then they trained a monocular depth estimation network end-to-end to conduct depth estimation from RGB images.

These methods demonstrate that focusness cues can greatly contribute to depth estimation, achieving superior results. However, there is still large room for further improvement in terms of spatial correlation excavation in the focal stack and multi-modal fusion between the RGB image and the focal stack. In this paper, we propose a deep learning based method that dynamically incorporates a RGB and a focal stack to confront these challenges.

\section{Method}
\subsection{The Overall Architecture}
   \begin{figure*}
\begin{center}
\includegraphics[width=1.0\linewidth]{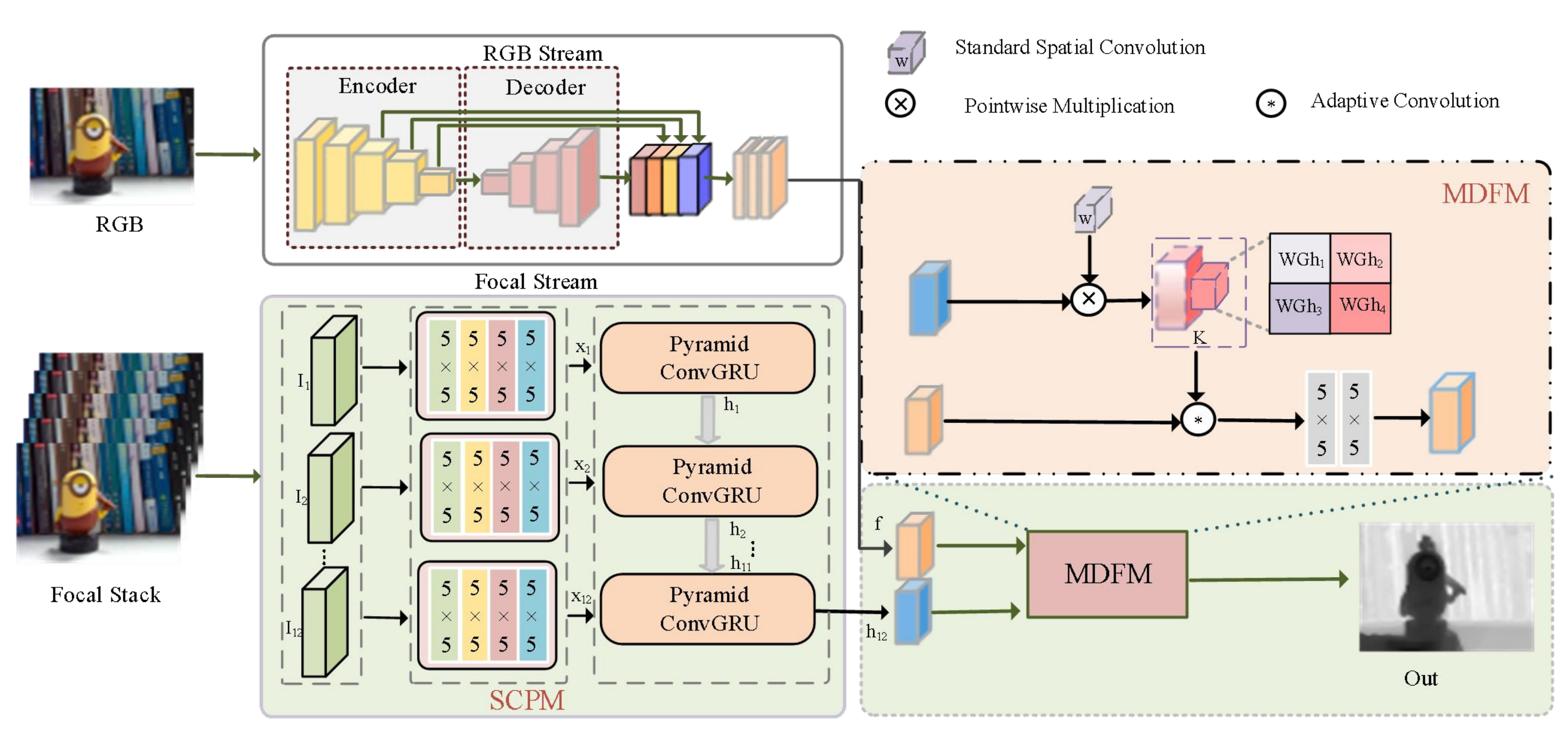}
\end{center}
\vspace{-0.55cm}
   \caption{The whole pipeline.}
\label{fig:long}
\vspace{-0.55cm}
\end{figure*}
In this paper, our goal is to excavate the spatial correlation  among focal slices and dynamically fuse RGB information and focusness features for focus-based depth estimation. The overall framework is described in Figure 1. \textbf{First}, the RGB image is fed into  the RGB stream which consists of a SeNet-154 network \cite{hu2018squeeze} pretrained on ImageNet \cite{deng2009imagenet}, a decoder employs four upsampling layer to gradually up-scale the final feature from the encoder and a refinement module  concatenate the feature from decoder and encoder in color channels  with three convolutional layers.  \textbf{Second}, all focal slices are fed into another focal stream to generate focusness information from focal slices, relying on the proposed spatial-correlation perception modelling (SCPM) designed to  excavate the spatial correlation in focal slices. \textbf{Last}, we propose a multi-modal dynamic fusion module (MDFM) to dynamically fuse multi-modal information  between RGB data and the focal stack in an adaptive way for providing good depth map. The details of our two modules will be discussed in the following two sections.

\subsection{Spatial-Correlation Perception Module (SCPM) }
Considering the stack of focal slices in a scene possess focus-area of multiple scales and focused at different depth, we aim to correlate the depth information with multi-scale focusness features. To do this, we propose a spatial-correlation perception module (SCPM), in which the spatial correlation between different focal slices is excavated by a proposed pyramid ConvGRU.  In this way, multi-scale focusness information in different slices can be transfered along the depth direction.

Specifically, we first feed 12 focal slices  $I_{1}(x, y, s, t)$, $I_{2}(x, y, s, t)$...$I_{n}(x, y, s, t)$ to four 5$\times$5 convolutional layers to encode focusness features $x_{i}$. This procedure can be defined as:
\begin{equation}
\begin{array}{l}
f_i \left( {I_i(x,y,s,t) ;\theta _i } \right) \to x_i,
 \end{array}
\end{equation}
where  $s \times t$ indicates the angular resolution, $x \times y$ indicates the spatial resolution and $i$ represents the i-th focus slice. $\theta$ denotes the parameters of the encoder layer and $f_{i}$ is a learning mapping function. Then, in order to excavate the spatial correlation between different focal slices, we draw ideas from recent works in \cite{eom2019temporally,yao2019lightweight,zhu2018towards}. Concretely, they use a typical ConvGRU to capture short and long term temporal dependencies. In our work, we consider the focusness features as a feature map sequence and design a pyramid ConvGRU. In order to pass multi-scale focusness features corresponding to focus area of multiple scales, the proposed pyramid ConvGRU use an atrous spatial pyramid pooling (ASPP) module \cite{chen2017deeplab} for each of the gates instead of convolution, which can encode multi-scale foucsness information by applying atrous convolution at multiple parallel filters with different rates and field-of-views. The dilation rates are 1, 3 and 5, respectively. The formulation of the proposed pyramid ConvGRU is:
\begin{equation}
r_i  = \sigma \left[ \begin{array}{l}
 W_{r1}  * [x_i ,h_{i - 1} ], \\
 W_{r3}  * [x_i ,h_{i - 1} ], \\
 W_{r5}  * [x_i ,h_{i - 1} ] \\
 \end{array} \right],
\end{equation}
\begin{equation}
z_i  = \sigma \left[ \begin{array}{l}
 W_{z1}  * [x_i ,h_{i - 1} ], \\
 W_{z3}  * [x_i ,h_{i - 1} ], \\
 W_{z5}  * [x_i ,h_{i - 1} ] \\
 \end{array} \right],
\end{equation}
\begin{equation}
\begin{array}{l}
n_i  = \tanh (x_i  * W_{xn }  + r_i \odot h_{i - 1}  * W_{hn }  + b_{n } ),
 \end{array}
\end{equation}
\begin{equation}
\begin{array}{l}
h_i  = \left( {1 - z_i } \right) \odot h_{i - 1}  + z_i  \odot n_i,
 \end{array}
\end{equation}
 \begin{wrapfigure}{r}{7.5cm}
\vspace{-0.4cm}
\centering
\includegraphics[width=0.99\linewidth] {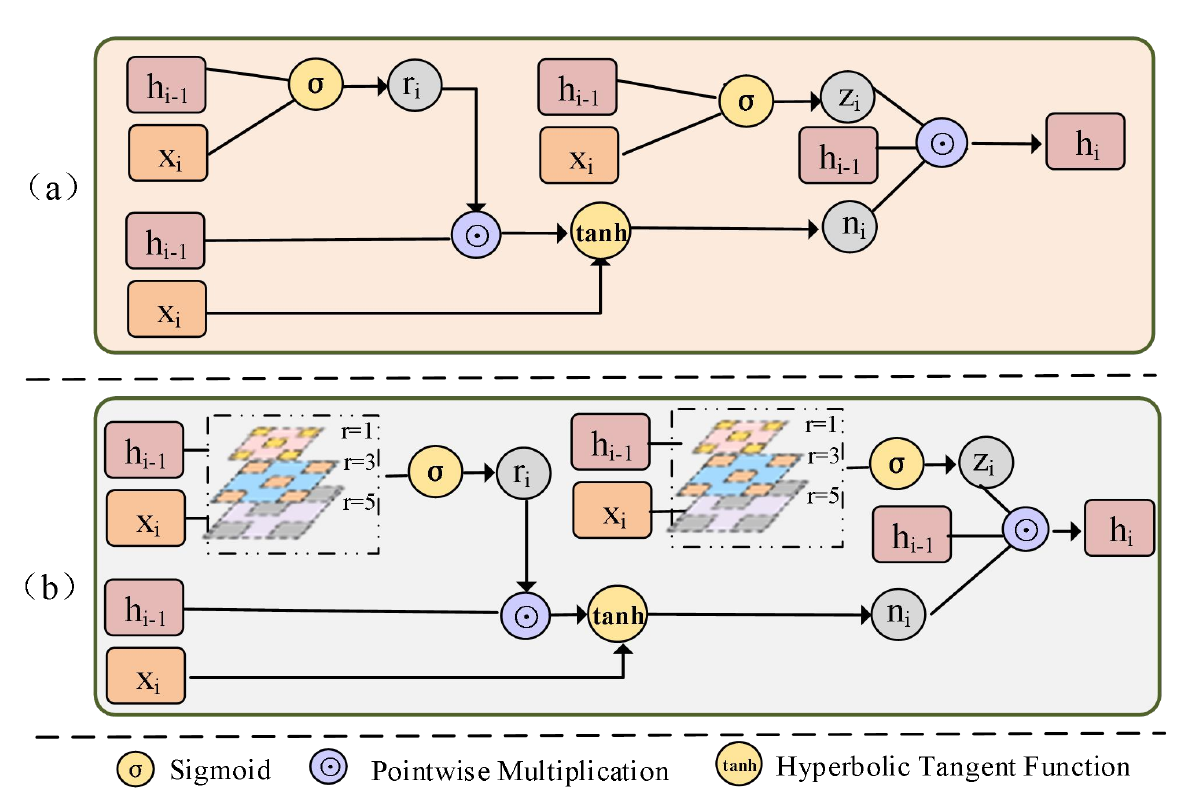}
\vspace{-0.6cm}
   \caption{ The architectures of ConvGRU and our pyramid ConvGRU. (a) ConvGRU; (b) pyramid ConvGRU. In ConvGRU, both  update and reset  gates are standard convolutional. In our pyramid ConvGRU, we use an atrous spatial pyramid pooling (ASPP) module  instead of convolution for each of the gates.}
\label{fig:short1}
\vspace{-0.45cm}
\end{wrapfigure}
 where all $W _{*}$ and $b _{*}$ are model parameters to be learned. $\sigma$ is sigmoid function. $\odot$ and $\ast$ are element-wise multiplication and convolution, respectively. The pyramid ConvGRU takes the i-th focusness feature $x_{i}$ and a previous output $h_{i-1}$ as input, outputs a new feature $h_{i}$ by combining $h_{i-1}$ and a candidate state $n_{i}$ weighted by an output of an update gate $z_{i}$. The update and reset gates, $z_{i}$ and $r_{i}$, selectively update multi-scale focusness information from the input focusness feature $x_{i}$ and the previous output $h_{i-1}$, automatically learn the spatial correlation from neighboring slices. Note that the updated focusness feature $h_{i}$ is not disappeared with information transfering, but passed to the next slice.

 \subsection{ Multi-Modal Dynamic Fusion Module (MDFM)}
 \begin{figure}[htp]
\begin{center}
\includegraphics[width=0.99\linewidth]{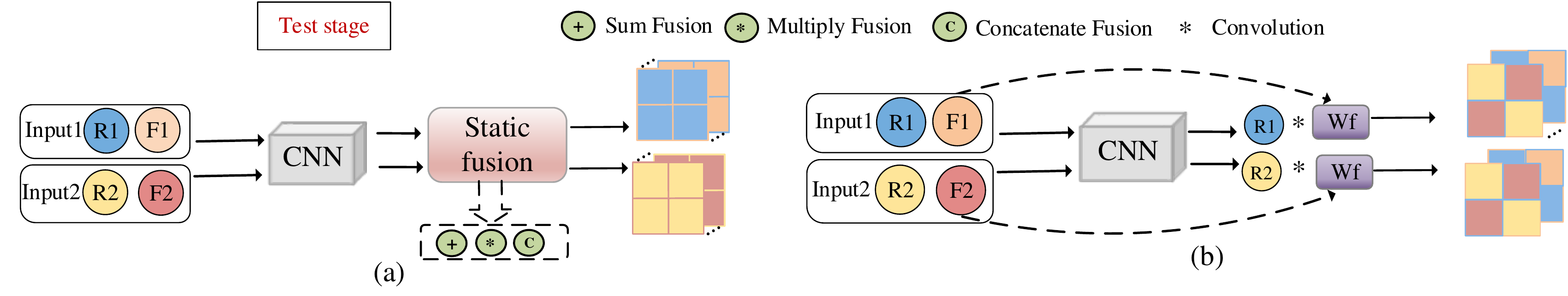}
\end{center}
\vspace{-4mm}
   \caption{ The architectures of static fusion and our fuse fusion. (a) Three different static fusion methods: sum fusion, weighted fusion, concatenate fusion.  (b) our MDFM.}
\label{fig:short1}
\vspace{-4mm}
\end{figure}
 As RGB images and focus slices imply different depth information, we consider that fusing different multi-modal information is important for depth estimation task. However, existing fusion schemes usually employ some manually set, including sum fusion, weighted  fusion, concatenate fusion. These static fusion schemes not suitable for our task because the RGB and focusness feature are not equivalent quantities, is prone to information loss. As shown in the Fig.4(a), these static fusion methods processes the entire image. When the network parameters are fixed, the convolution kernel does not change with the input features, ignoring the relationship between multi-modal features. Therefore, a more proper and effective strategy should be considered. To this end, we introduce a multi-modal dynamic fusion module (MDFM) to dynamically fuse the RGB features and focusness features in an adaptive manner. In this module, the filter varies with focusness features is used to convolve with  RGB information , thereby avoiding information  loss.
 \begin{wrapfigure}{r}{7.5cm}
\vspace{-0.4cm}
\centering
\includegraphics[width=0.99\linewidth] {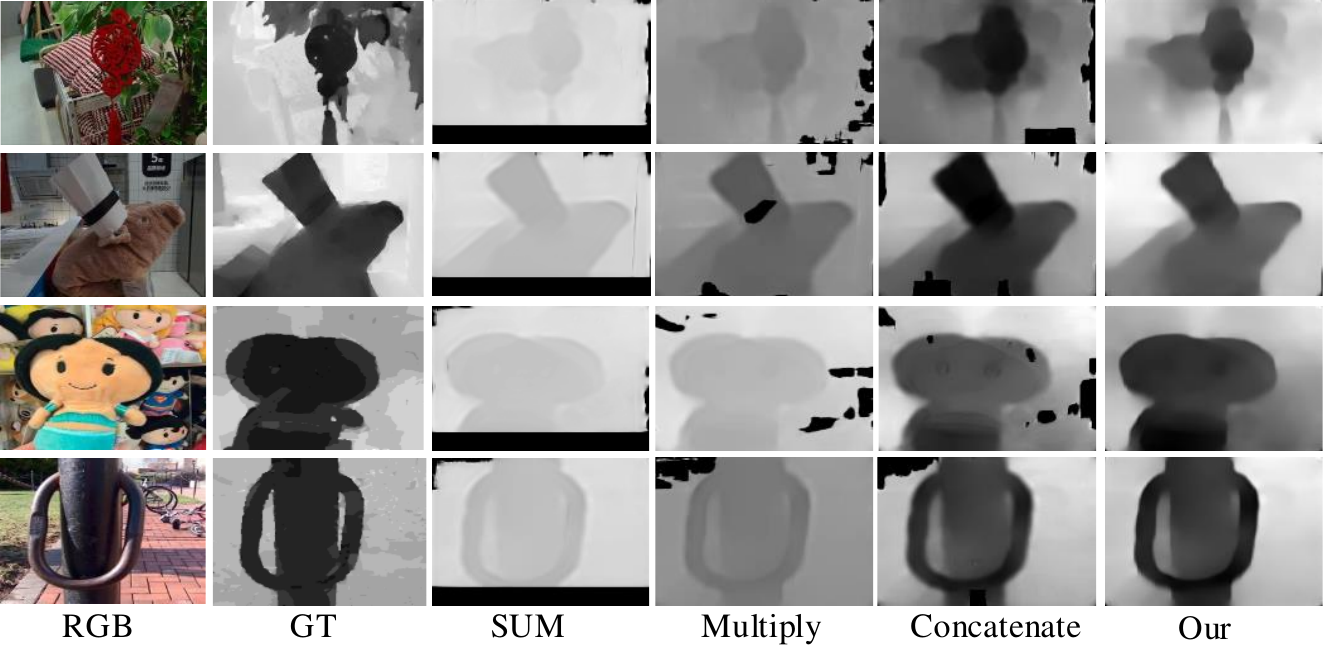}
\vspace{-0.6cm}
    \caption{Visual results of the effect of MDFM.  $Concatenate$ represents Concatenate Fusion. $SUM$ represents Sum Fusion. $Weight$ represents Weighted Fusion. }
\label{fig:short1}
\vspace{-0.45cm}
\end{wrapfigure}

 Specifically, the process of our module consists of two steps: 1) we choose a adaptive kernel [33] instead of the spatially invariant convolution. In our work, the standard spatial convolution W is adapted at each pixel using focuesness features f via the adaptive kernel. Therefore, when the focusness features change, the parameters of the convolution kernel also change;  2) we apply the generated adaptive convolution kernel  to RGB features f, making the whole network dynamically fuse multi-modal information to get an accurate depth map.

 For the output from the spatial-correlation perception module h, we have:
\begin{equation}
\begin{array}{l}
d_i  = \sum\limits_{j \in \Omega (i)} {\exp ( - \frac{1}{2}(h_i  - h_j )} ^T (h_i  - h_j ))W\left[ {p_i  - p_j } \right]f_j  + b
 \end{array}
\end{equation}
where $[pi-pj]$ is index of the spatial dimensions of an array with 2D spatial offsets, i and j represent the pixel coordinates, $W$ is a standard spatial convolution W, $f_{j}$ represents the output from the RGB stream, $b$ is bias term, $d$ represents the final depth prediction map. Before we perform filtering operation, the filter has a pre-defined form and depend on the content of focusness features. It is hoped that the prediction map depends on both the RGB features and the reliable focusness information. Finally, we refine the depth map by adding two consecutive convolutional layers.  As shown in the Fig.4(b), in test stage, compared to traditional methods of fusing multi-modal information which parameters do not change with input features, our model dynamically selects filter parameters for convolution based on the content of the focusness features and adaptively selects multimodal features for fusion, avoiding information loss.

 \subsection{Loss Functions}
 In this section, we calculate the reconstruction error between the predicted images and the ground truth. We define the loss by:
\begin{equation}
\begin{array}{l}
L = l_{depth}  + \lambda l_{grad}  + \mu l_{normal}
 \end{array}
\end{equation}
where $\lambda$, $\mu$ are weighting coefficients. Here we set as $\lambda =1$, $\mu = 1$. We use a logarithm of depth errors as $l_{depth}$ to minimize the difference between the depth estimate map $d_{i}$ and its ground truth $g_{i}$:
\begin{equation}
\begin{array}{l}
l_{depth}  = \frac{1}{n}\sum\limits_{i = 1}^n {\ln \left( {\left\| {d_i  - g_i } \right\|_1  + \alpha } \right)}
 \end{array}
\end{equation}
where $\alpha(> 0)$ is a parameter we set $\alpha = 0.5$. In order to deal with edge distortion problems caused by CNNs training, we consider the following loss function of the gradients of depth:
\begin{equation}
\begin{array}{l}
l_{grad}  = \frac{1}{n}\sum\limits_{i = 1}^n {\left( {F\left( {\nabla _x \left( {\left\| {d_i  - g_i } \right\|_1 } \right)} \right) + F\left( {\nabla _y \left( {\left\| {d_i  - g_i } \right\|_1 } \right)} \right)} \right)}
 \end{array}
\end{equation}
where $\bigtriangledown x (*) $ is the spatial derivative of $\|d_{i}-g_{i}\|$  computed at the ith pixel with respect to $ x$, and so on. To deal with such small depth structures and further improve fine details of depth maps, we consider yet another loss for training, which measures accuracy of the normal to the surface of an estimated depth map with respect to its ground truth:
\begin{equation}
\begin{array}{l}
l_{normal}  = \frac{1}{n}\sum\limits_{i = 1}^n {\left( {1 - \frac{{\left\langle {n_i^d ,n_i^g } \right\rangle }}{{\sqrt {\left\langle {n_i^d ,n_i^d } \right\rangle } ,\sqrt {\left\langle {n_i^g ,n_i^g } \right\rangle } }}} \right)}
 \end{array}
\end{equation}
where $< ·,· > $ denotes the inner product of vectors. Denoting the surface normal of an estimated depth map and its ground truth by $n_i^d  = \left[ { - \nabla _x \left( {d_{_i } } \right),{\rm{ }} - \nabla _y \left( {d_{_i } } \right),1} \right]^{^T }$, $n_i^g  = \left[ { - \nabla _x \left( {g_{_i } } \right),{\rm{ }} - \nabla _y \left( {g_{_i } } \right),1} \right]^{^T }$

\section{Experiments }
\subsection{Experiments Setup}
\subsubsection{Datasets.}
To demonstrate the effectiveness of the proposed approach, we conduct experiments on DUT-LFDD dataset, LFSD dataset and a Mobile Phone dataset. Note that although there are some synthetic
light field (LF) datasets proovided by [12] which consist of multiview images and corresponding depth maps. The models trained on such synthetic data have trouble generalizing to real-world data. Furthermore, the focal stack can not be correctly synthesized from the multi-view images provided in HCI, because the changes between views are not transitional but rotational.

\noindent{\textbf{DUT-LFDD: }This dataset contains 967 real-world light-field samples coming from a large variety of indoor and outdoor scenes. We randomly select 630 samples for training and the remaining 337 samples for testing. All images are captured by commercial Lytro Illum camera in real life scenes and contains many challenging scenes. Each light field sample consists of a RGB image, a focal stack with 12 focal slices focusing at different depths and a depth image corresponding to the RGB image.

\noindent{\textbf{LFSD: } The dataset proposed by \cite{li2014saliency} contains 100 light fields captured by Lytro camera, including 60 indoor and 40 outdoor scenes. Each light field consists of an RGB image, focal slices and depth map.

\noindent{\textbf{Mobile Phone dataset: } The Mobile Phone dataset provided by previous researchers \cite{suwajanakorn2015depth}. They continuously captured images of size  640$\times$360 pixels using a Samsung Galaxy S3 phone during  auto-focusing. The dataset provides focal stack and RGB image of different scenes (number of frames in parenthesis): plants(23), bottles(31), fruits(30), metals(33), window(27), telephone(33), etc. For each scene, we choose 12 focal slices and a RGB image to evaluate our model.

To prevent the overfitting problem, we augment the training set by the following operations:
\begin{itemize}
\item Flipping: we only consider horizontal flipping (i.e. mirroring) of images at a probability of 0.5.
\item Rotating: the RGB image, focal stack and the depth image are rotated by a random degree r $\in$ [-5,5].
\item Color Jitter: brightness, contrast, and saturation values of the sample are randomly scaled by c $\in$ [0.6,1.4].
\end{itemize}
\subsubsection{Evaluation Metrics.}
We adopt seven metrics for comprehensive evaluation, including Mean Eelation error(Abs Rel), Squared Relative error (Sq Rel), Root Mean Squared error (RMSE), Mean log 10 error (RMSE log), Accuracy with threshold ($thr = 1.25, 1.25^{2}, 1.25^{3}]$). They are universally-agreed and standard for evaluating a depth estimation model and well explained in many literatures \cite{zhao2019geometry,tosi2019learning,zheng2018t2net}.
\begin{table*}[h]
\vspace{-0.1cm}
  \centering
  \setlength{\tabcolsep}{2.0mm}
  \begin{threeparttable}
  \caption{ Quantitative results of the ablation analysis for our SCPM.}
  \label{tab:performance_comparison}
    \begin{tabular}{cccccccc}%cc}p{0.8cm}<{\centering}p{0.8cm}<{\centering}p{0.8cm}<{\centering}p{0.8cm}<{\centering}p{0.6cm}<{\centering}p{0.6cm}<{\centering}p{0.6cm}<{\centering}}
    \toprule
    \multicolumn{1}{c}{\multirow{2}{*}{\scriptsize{Methods}}}&
    \multicolumn{4}{c}{\scriptsize Error metric$\downarrow$}&\multicolumn{3}{c}{\scriptsize Accuracy metric$\uparrow$}\cr
    \cmidrule(lr){2-5} \cmidrule(lr){6-8}
    \multirow{2}{*}{}&\scriptsize{RMSE}&\scriptsize{RMSE Log}&\scriptsize{Abs Rel}&\scriptsize{Sq Rel}&\scriptsize{$\delta<1.25$}&\scriptsize{$\delta<1.25^{2}$}&\scriptsize{$\delta<1.25^{3}$}\cr
     \midrule
        \multirow{1}{*} {\scriptsize {2DCNN}}&\scriptsize0.3667&\scriptsize0.0501&\scriptsize0.1879&\scriptsize0.1015&\scriptsize{0.7102}&\scriptsize{0.9435}&\scriptsize{0.9930}\cr
    \multirow{1}{*} {\scriptsize {GRU}}&\scriptsize0.3648&\scriptsize0.0492&\scriptsize0.1821&\scriptsize0.0976&\scriptsize0.7159&\scriptsize0.9444&\scriptsize0.9931\cr
            \multirow{1}{*} {\scriptsize {SCPM}}&\scriptsize0.3457&\scriptsize0.0453&\scriptsize0.1707&\scriptsize0.0864&\scriptsize0.7342&\scriptsize0.9516&\scriptsize0.9952\cr
    \bottomrule
    \end{tabular}
    \end{threeparttable}
    \vspace{-0.35cm}
\end{table*}
\begin{itemize}
\item Mean relation error (Abs Rel):
$\frac{1}{T}\sum\nolimits_{i \in T} {\frac{{\left| {d_i  - d_i^{gt} } \right|}}{{d_i^{gt} }}}$
\item Root mean squared error (RMSE):

${\frac{1}{T}\sum\nolimits_{i \in T} {\left\| {d_i  - d_i^{gt} } \right\|} ^2 }$
\item Mean log 10 error (RMSE log):

${\sqrt {\frac{1}{T}\sum\nolimits_{i \in T} {\left\| {\log (d_i ) - \log (d_i^{gt} )} \right\|} ^2 }}$
\item Squared relative error (Sq Rel): $\frac{1}{T}\sum\nolimits_{i \in T} {\frac{{\left\| {d_i  - d_i^{gt} } \right\|^2 }}{{d_i^{gt} }}}$
\item Accuracy with threshold ($\delta$):$\max \left( {\frac{{d_i }}{{d_i^{gt} }},\frac{{d_i^{gt} }}{{d_i }}} \right) = \delta  < thr $
\end{itemize}

\subsubsection{Implementation Details.}
We implement our network based on Pytorch framework with one Nvidia GTX 1080Ti GPU. We train the RGB stream and focal stream using Adam optimizer with an initial learning rate of 0.0001, and reduce it to 10\% for every 5 epochs. We set $\beta1$ = 0.9, $\beta2$ = 0.999, and use weight decay of 0.0001. The encoder module in the RGB stream is initialized by a model pretained with the ImageNet dataset. The other layers in the network are randomly initialized.  The batchsize is 1 and maximum epoch is set 80.
 \begin{figure}[htp]
\begin{center}
\includegraphics[width=0.99\linewidth]{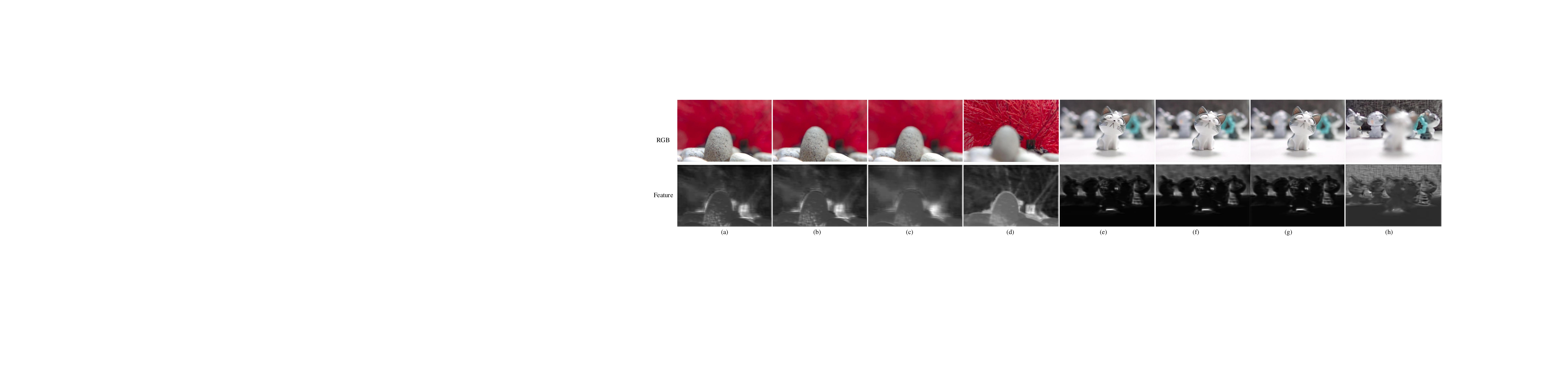}
\end{center}
\vspace{-4mm}
   \caption{ The architectures of static fusion and our fuse fusion. (a) Three different static fusion methods: sum fusion, weighted fusion, concatenate fusion.  (b) our MDFM.}
\label{fig:short1}
\vspace{-4mm}
\end{figure}

\subsection{Ablation Studies}

\subsubsection{Effect of SCPM.}The SCPM is proposed to  excavate the spatial correlation between different focal slices. To verify the effectiveness of the SCPM, we first replace the SCPM with 7 convolution layers (noted as $2DCNN$). Table 1 shows that the SCPM improves the RMSE performances by nearly 2\% than 2DCNN. We believe that this improvement is due to the cyclic structure retaining the spatial relationship between different focal features. Then we compare the performance of our pyramid ConvGRU and ConvGRU. We replace pyramid ConvGRU with the ConvGRU. And we can observe from Table 1 that the improvements in performance are achieved by using our pyramid ConvGRU. These improvements are resonable since  our model can adapt to different  sizes of focus area compared to ConvGRU. Furthermore, we randomly select four different focus slices for visualization after inputting them into pyramid ConvGRU in Fig.5. (a)-(c) and (e)-(g) represent the first three of the 12 focus slices of each scene. (d) and (h) represent the output of the last pyramid ConvGRU in each scene.  We can see from Fig.5 that
the feature maps of different focus slices contain different details and are passed by our pyramid ConvGRU.
\begin{table}[htbp]
  \centering
  \setlength{\tabcolsep}{2.0mm}
  \begin{threeparttable}
  \caption{ Quantitative results of the ablation analysis for our MDFM. $Concatenate$ represents Concatenate Fusion. $SUM$ represents Sum Fusion. $Weight$ represents Weighted Fusion.}
  \label{tab:performance_comparison}
    \begin{tabular}{ccp{0.65cm}<{\centering}p{0.65cm}<{\centering}p{0.65cm}<{\centering}p{0.65cm}<{\centering}p{0.65cm}<{\centering}p{0.65cm}<{\centering}p{0.65cm}<{\centering}}
    \toprule
    \multicolumn{1}{c}{\multirow{2}{*}{Type}}&
    \multicolumn{1}{c}{\multirow{2}{*}{Methods}}&
    \multicolumn{4}{c}{Error metric}&\multicolumn{3}{c}{Accuracy metric}\cr
    \cmidrule(lr){3-6} \cmidrule(lr){7-9}
     &{}&\footnotesize{RMSE}&\footnotesize{RMSE Log}&\footnotesize{Abs Rel}&\footnotesize{Sq Rel}&\footnotesize{$\delta<1.25$}&\footnotesize{$\delta<1.25^{2}$}&\footnotesize{$\delta<1.25^{3}$} \cr
    \midrule
        \multirow{4}{*}{Our dataset}& {\scriptsize{SUM}}&\scriptsize0.6600&\scriptsize0.1564&\scriptsize0.3777&\scriptsize0.3211&\scriptsize{0.4186}&\scriptsize{0.6838}&\scriptsize{0.9184}\cr
  &\scriptsize {Weight}& \scriptsize{0.4877}&\scriptsize{0.0921}&\scriptsize{0.2603}&\scriptsize {0.1652}&\scriptsize{0.5190}&\scriptsize{0.8584}&\scriptsize{0.9749}\cr
 & {\scriptsize {Concatenate}}&\scriptsize{0.4268}&\scriptsize{0.0705}&\scriptsize{0.1917}&\scriptsize0.1065&\scriptsize{0.6250}&\scriptsize{0.9122}&\scriptsize{0.9790}\cr
 &{\scriptsize {our}}&\scriptsize0.3457&\scriptsize0.0453&\scriptsize0.1707&\scriptsize0.0864&\scriptsize0.7342&\scriptsize0.9516&\scriptsize0.9952\cr
  \midrule
     \multirow{4}{*}{LFSD}& {\scriptsize {SUM}}&\scriptsize0.7037&\scriptsize0.1788&\scriptsize0.3977&\scriptsize0.3484&\scriptsize{0.3971}&\scriptsize{0.6325}&\scriptsize{0.8889}\cr
  & {\scriptsize {Weight}}& \scriptsize{0.4667}&\scriptsize{0.0829}&\scriptsize {0.2422 }&\scriptsize { 0.1501}&\scriptsize{ 0.5777}&\scriptsize{ 0.8823 }&\scriptsize{ 0.9756}\cr
  &{\scriptsize {Concatenate}}&\scriptsize{0.4284}&\scriptsize{ 0.0697}&\scriptsize{0.1858}&\scriptsize{0.1042}&\scriptsize{0.6570}&\scriptsize{0.9081}&\scriptsize{0.9752}\cr
& {\scriptsize {our}}&\scriptsize{0.3612}&\scriptsize{0.0550}&{\scriptsize{0.1796}}&{\scriptsize{0.0901}}&{\scriptsize{0.6973}}&{\scriptsize{0.9339}}&{\scriptsize{0.9874}}\cr
 \midrule
    \end{tabular}
    \end{threeparttable}
\end{table}

\subsubsection{Effect of MDFM.}
The MDFM is proposed for dynamically fusing multi-modal information between RGB features and focusness features. To demonstrate the effectiveness of the MDFM, we compare the MDFM with a variety of conventional fusion methods, including Concatenate Fusion(noted as $Concatenate$), Sum Fusion(noted as $SUM$), Weighted Fusion(noted as $Weight$). For better comparison, we replace the fusion block in the framework. The results of the comparison are shown in Figure 4. It can be seen that the quality of depth map achieves accumulative improvements by a large margin with the MDFM. Especially in regions of depth discontinuities, the MDFM is able to recover the depth more accurately than the concatenate fusion and preserves more structural information compared to other static fusion methods. Numerically, our proposed MDFM reduces the RMSE performances by nearly 8\% than concatenate fusion and as shown in Table 2.

\begin{table}[htbp]
  \centering
  \setlength{\tabcolsep}{2.5mm}
  \begin{threeparttable}
  \caption{Quantitative comparison with state-of-the-art methods. From top to bottom: our dataset, LFSD dataset. $*$ respects non-deep-learning methods. The best result results are shown in \textbf{boldface}. }
  \label{tab:performance_comparison}
    \begin{tabular}{ccp{0.65cm}<{\centering}p{0.65cm}<{\centering}p{0.65cm}<{\centering}p{0.65cm}<{\centering}p{0.65cm}<{\centering}p{0.65cm}<{\centering}p{0.65cm}<{\centering}}
    \toprule
    \multicolumn{1}{c}{\multirow{2}{*}{Type}}&
    \multicolumn{1}{c}{\multirow{2}{*}{Methods}}&
    \multicolumn{4}{c}{Error metric}&\multicolumn{3}{c}{Accuracy metric}\cr
    \cmidrule(lr){3-6} \cmidrule(lr){7-9}
     &{}&\footnotesize{RMSE}&\footnotesize{RMSE Log}&\footnotesize{Abs Rel}&\footnotesize{Sq Rel}&\footnotesize{$\delta<1.25$}&\footnotesize{$\delta<1.25^{2}$}&\footnotesize{$\delta<1.25^{3}$} \cr
    \midrule
     \multirow{6}{*}{Our dataset}&Our&{\bfseries{0.3457}}&{\bfseries{0.0453}}&{\bfseries{0.1707}}&{\bfseries{0.0864}}&{\bfseries{0.7342}}&{\bfseries{0.9516}}&{\bfseries{0.9952}}\cr
    &DDFF &0.5255&0.1042&0.2666&0.1834&0.4944&0.8202&0.9667\cr
    &EPINET &0.4974&0.0959&0.2324&0.1434&0.5010&0.8375&0.9837\cr
    &VDFF$^*$&0.7192&0.1887&0.3887&0.3808&0.4040&0.6593&0.8505\cr
    &PADMM$^*$&0.4730&0.0912&0.2253&0.1509&0.5891&0.8560&0.9577\cr
    &LF$^*$&0.6897&0.1436&0.3835&0.3790&0.4913&0.7549&0.8783\cr
    &LF$\_$OCC$^*$&0.6233&0.1432&0.3109&0.2510&0.4524&0.7464&0.9127\cr
    \midrule
     \multirow{6}{*}{LFSD}&Our&\bfseries{0.3612}&\bfseries{0.0550}&{\bfseries{0.1796}}&{\bfseries{0.0901}}&{\bfseries{0.6973}}&{\bfseries{0.9339}}&{\bfseries{0.9874}}\cr
    &DDFF&0.4255&0.0717&0.2128&0.1204&0.6185&0.8916&0.9860\cr
    &VDFF$^*$&0.5747&0.1258&0.3320&0.2660&0.4730&0.7823&0.9359\cr
    &PADMM$^*$&{0.4238}&{0.0722}&0.2153&0.1336&0.6536&0.8880&0.9770\cr
     &LF$^*$&-&-&-&-&-&-&-\cr
    &LF$\_$OCC$^*$&-&-&-&-&-&-&-\cr
    &EPINET &-&-&-&-&-&-&-\cr
    \midrule
    \end{tabular}
    \end{threeparttable}
\vspace{-6mm}
\end{table}
\subsection{Comparison with State-of-the-arts}
We compare our method with 6 other state-of-the-arts light field depth estimation method, containing both deep-learning-based methods and traditional methods (marked with $*$). Deep-learning-based method is DDFF \cite{hazirbas2018deep}, EPINET \cite{shin2018epinet}, traditional methods are VDFF$^*$ \cite{moeller2015variational}, PADMM$^*$ \cite{javidnia2018application}, LF$^*$\cite{jeon2015accurate}, LF$\_$OCC$^*$ \cite{wang2015occlusion}. For fair comparisons,  the results from competing methods are generated by authorized codes. To verify the generalization and applicabilityof our network, we conduct experiments on three dataset.
\begin{figure}[htpb]
\begin{center}
\includegraphics[width=1.0\linewidth]{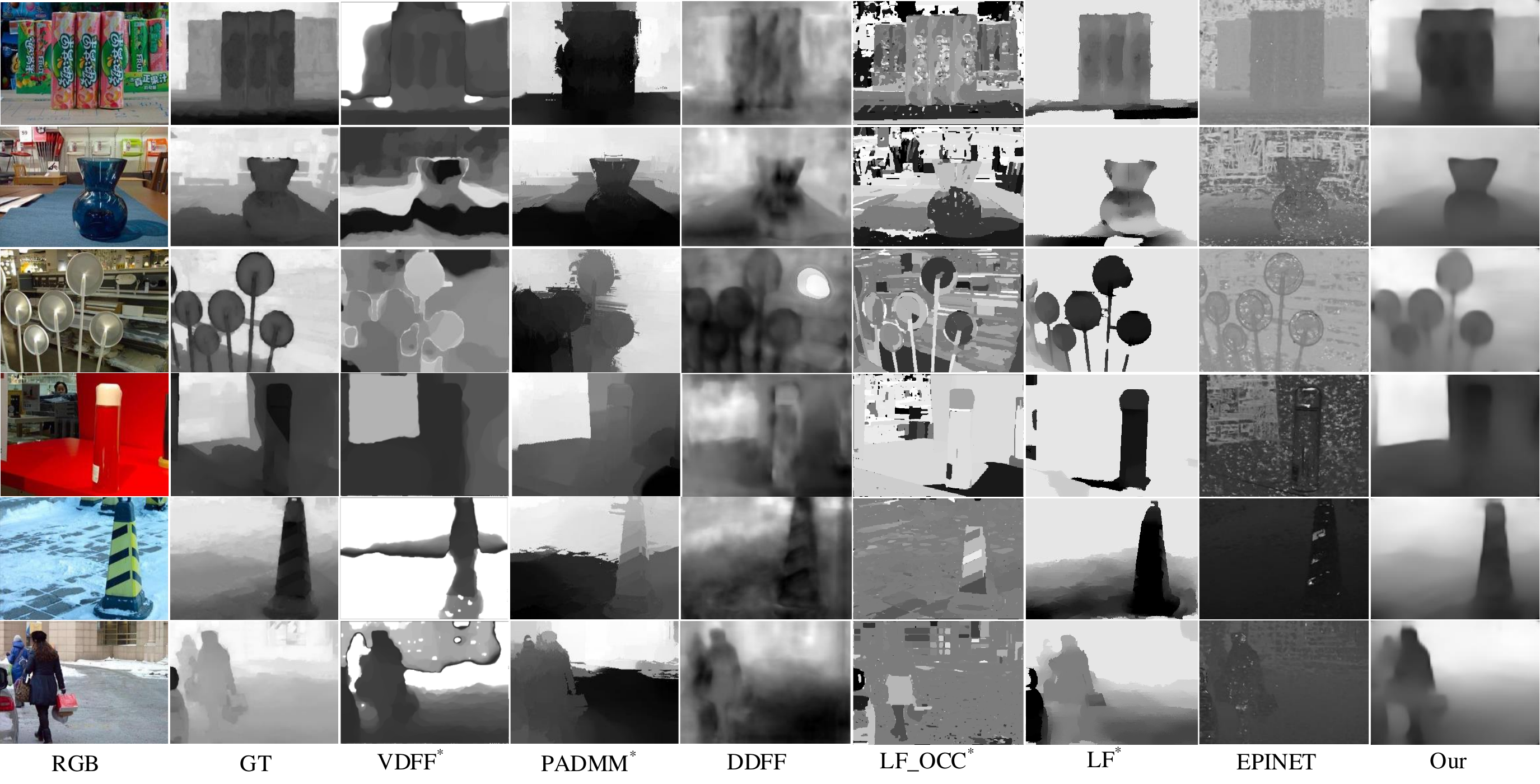}
\end{center}
\vspace{-4mm}
   \caption{Visual comparisons of our method over some other approaches. Our method produces more accurate depth maps than others, and our results are more consistent with the ground truths (denoted as ’GT’).}
\label{fig:long}
\vspace{-4mm}
\end{figure}
\subsubsection{Quantitative Evaluation.}
 As shown in Table 3, our method is able to clearly outperform the other state-of-the-art methods on our dataset in terms of seven evaluation metrics. Not only that, we also apply the model parameters trained on our dataset directly to the LFSD dataset for testing, and our method achieves significant advantages, such as Top-1 accuracies (Sq Rel) and Top-1 accuracies (RMSE). Note that the LFSD dataset only provides the focal slices, RGB images and depth maps, therefore, we only compare our method with 3 focus-based methods on LFSD dataset.

\subsubsection{Qualitative Evaluation.}
Figure 5 provides some challenging samples of results comparing our method with other state-of-the-art methods. It can be seen that our method can achieve accurate prediction, when foreground and background are similar as shown in the $1^{th}$ and $4 ^{th}$ rows, when some areas are textureless as shown in the $3^{th}$ row, when smooth surface as shown in the $2^{th}$ row.

We further illustrate the visual results of our method on LFSD dataset in Figure 6. From the $5^{th}$ line and the $6^{th}$ , we can find that compared with traditional methods, our network retains more detailed information and better maintains the structure of the depth map. From the $7^{th}$ line, we can find the results of DDFF contain a lot of noise. This is beacuse DDFF used a standard 2D CNN to learns filters that extend across the entire focal stack. In contrast, our model preserves the spatial correlation between focal slices and effectively fuses the focusness feature and RGB information to reduce information loss.
 \begin{figure}[!h]
\begin{center}
\includegraphics[width=0.99\linewidth]{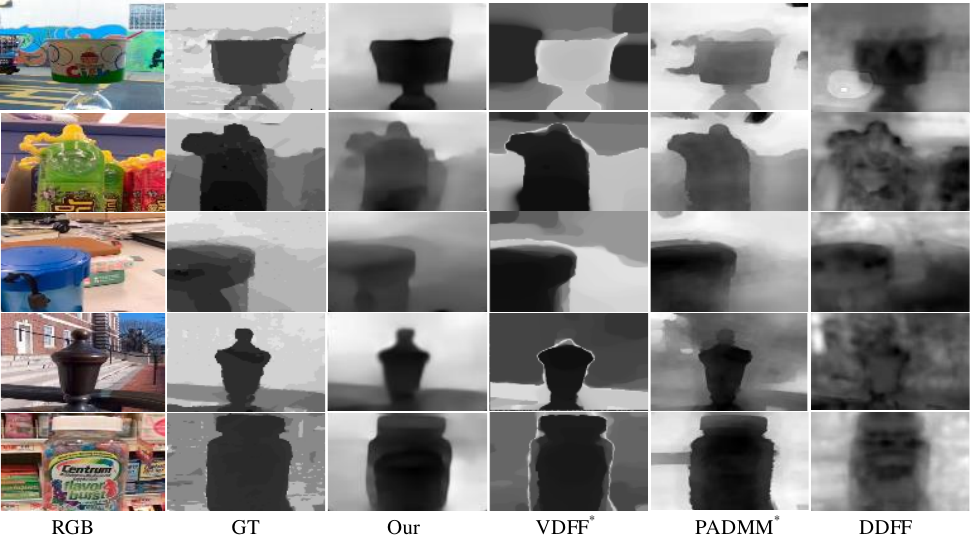}
\end{center}
\vspace{-4mm}
   \caption{ Comparison with deep learning-based methods on LFSD dataset: RGB images, the corresponding ground truth, our estimated depth maps and other state-of-the-arts light field depth estimation results.}
\label{fig:short2}
\vspace{-4mm}
\end{figure}
\subsection{Mobile Phone Results}
The common consumer level cameras also can capture focal stack. To expand the practical application of our network, we test our network on the focal stack captured captured by a Samsung Galaxy S3 phone, provided by the authors of  \cite{suwajanakorn2015depth}. For ease of testing, we choose 12 focal slices and a RGB image from each scene.  Figure 9 provides a qualitative comparison of different deep learning-based methods. It can be seen that the proposed method allows to recover high quality depth maps with less noise and sharper object boundaries, although the model was not trained on this specific dataset.  The impressive results show that the proposed method can be used for further applications in our daily life. Further, our model is not limited by the light field camera because the smart-phone camera can capture focal stacks by manipulating the focus distance programmatically and it represents the dominant modality of image capture.

 \begin{figure}[htb]
\begin{center}
\includegraphics[width=0.99\linewidth]{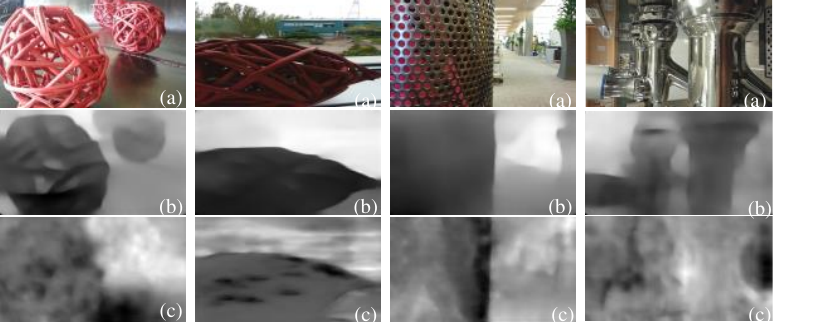}
\end{center}
\vspace{-4mm}
   \caption{ Comparison with deep learning-based methods on Mobile Phone dataset:(a) RGB images (b) our results (c) DDFF\cite{hazirbas2018deep}}
\label{fig:short1}
\vspace{-4mm}
\end{figure}
\section{Conclusion}
In this paper, we propose a multimodal learning which incorporates RGB data and the focal stack for focus-based depth estimation. Our SCPM excavates the spatial correlation between different focal slices and sequently pass multi-scale focusness information along the depth direction by  exploiting our proposed pyramid ConvGRU. The MDFM dynamically fuse the RGB features and focusness features in an adaptive manner. In this module, the filter varies with focusness features is used to convolve with RGB information. Our experiments show that the proposed method achieves superior performance, especially in challenging scenes.  Our extensive evaluation shows that the proposed network can be applied on common consumer level cameras data successfully.
\bibliographystyle{splncs04}
\bibliography{egbib}
\end{document}